\documentclass{article}
\usepackage{spconf,amsmath,graphicx}

\usepackage{enumitem}
\setlist{nosep, leftmargin=14pt}

\usepackage{amssymb}
\usepackage{url}
\def\eopp{{EOpp}}
\def\eodd{{EOdds}}

\title{Explanation-Aware Fairness Beyond Output Parity: Joint Output-Attribution Regularization for Medical Image Classification}
\name{G{\"o}khan~{\"O}zbulak$^{\star \dagger}$ \qquad Oscar~Jimenez-del-Toro$^{\star}$ \qquad Lilian~Berton$^{\ddagger}$ \qquad Andr{\'e}~Anjos$^{\star}$}

\address{$^{\star}$ Idiap Research Institute, Martigny, Switzerland \\
     $^{\dagger}$ \'{E}cole Polytechnique F\'{e}d\'{e}rale de Lausanne (EPFL), Lausanne, Switzerland \\
     $^{\ddagger}$ Universidade Federal de S{\~a}o Paulo (UNIFESP), S{\~a}o Paulo, Brazil}

\begin{document}
%\ninept
%
\maketitle
\begin{abstract}
Output-level fairness does not constrain whether demographic groups rely on similar predictive contributions in representation space.
For a nonlinear encoder with a linear head, we show that outcome-conditioned output score disparity equals the signed sum of feature attribution differences across subgroups by allowing small prediction gaps to arise through cancellation effect.
We test whether an attribution disparity penalty complements output-level equalized odds based fairness surrogates.
Across 960 DINOv2-LoRA models on HAM10000 and Harvard-GF datasets for gender and race fairness, attribution augmentation increases utility-fairness trade-off performance for hypervolume in six different settings with paired 95\% confidence intervals and relative gains of 1.49-10.57\%.
Under a predefined utility constraint, augmented model families also achieve 9.0-38.7\% lower equalized odds difference, lower attribution disparity, and reduced cancellation ratio while preserving predictive utility.
These results support attribution fairness as a complementary explanation-aware constraint on output fairness.
%Source code is available for reproducibility at \url{https://filesender.switch.ch/filesender2/?s=download&token=6e8e15a9-068e-47e5-84d6-dfcd897116b7}.
\end{abstract}
\section{Introduction}
\label{sec:intro}

Fairness is a central requirement for medical imaging artificial intelligence (AI) because predictive performance can vary systematically across demographic groups.
Prior work has reported gender and race based disparities in computer-aided diagnosis \cite{larrazabal2020gender,seyyed2021underdiagnosis}, while medical images can encode demographic information that models may exploit as predictive shortcuts \cite{yang2024limits}.
These observations motivate group-fairness criteria such as equal opportunity (\eopp) and equalized odds (\eodd) to quantify disparities in model predictions and errors across sensitive groups \cite{hardt2016equality}.
Fairness mitigation is nevertheless multi-objective: different regularization strengths induce distinct utility-fairness operating points, so evaluating a method at a single selected point can mask differences in the quality of its attainable trade-off.
Pareto frontiers (PF) and hypervolume (HV) based analyses provide a natural family-level characterization of such solution sets \cite{ozbulak_multi-objective_2025}.

Output-level fairness does not explicitly characterize how task-relevant predictive feature contributions differ across demographic groups.
Explainability provides a complementary perspective by exposing how model outputs are composed from individual contributions.
Feature-attribution methods have been used to decompose fairness measures \cite{lundberg2020explaining}, characterize procedural fairness \cite{wang2026procedural}, and support bias-aware learning in healthcare \cite{shahbazian2025joint}.
However, the mathematical relationship between group disparities in predictive feature contributions and the resulting classifier output gap remains insufficiently characterized.

We show that, for a linear classification head, outcome-conditioned output score disparity equals the signed sum of attribution differences across subgroups.
Consequently, opposing group differences can cancel during aggregation by yielding a small output gap despite substantial disparity in the predictive evidence.
Our central research question is therefore: \textit{Can attribution-level fairness provide complementary information to output-only regularization such that incorporating the same attribution penalty into different equalized odds surrogates improves their attainable utility-fairness trade-offs?}
We hypothesize that \textit{attribution augmentation yields more favorable utility-fairness Pareto frontiers than the corresponding output-only regularizers.}
We test this hypothesis across two distinct medical imaging modalities and sensitive-attribute settings as skin-lesion classification with gender and glaucoma detection with race.

\section{Explanation-Aware Fairness}
\label{sec:method}

We formulate explanation-aware fairness by distinguishing model output disparities from differences in the task-relevant feature contributions underlying predictions.
Motivated by cancellation of opposing contribution differences at the prediction level, we first establish the theoretical relationship between these disparities under \eodd{} conditioning.
We then define an attribution-level fairness objective and combine it with output-level regularization in a joint training formulation.

\subsection{Output-Attribution Disparity}
\label{ssec:out_attr}

Let $x$ denote an input image and $z=E_{\theta}(x) \in \mathbb{R}^{D}$ the $D$-dimensional representation produced by an arbitrary nonlinear encoder.
A linear binary classification head produces
\begin{equation}
    s(x)=w^\top z + b
    \label{eq:raw_score}
\end{equation}
We quantify the contribution of feature dimension $k$ by
\begin{align}
    \phi_{k}(x)&=z_{k}\frac{\partial s(x)}{\partial z_{k}}=w_{k}z_{k} 
    \label{eq:attr} \\
    s(x)-b&=\sum_{k=1}^{D}\phi_{k}(x)
    \label{eq:completeness}
\end{align}

For sensitive attribute $A$, outcome $Y$, and outcome value $y$, we define
\begin{align}
    \Delta^{out}_{a,y}&=\mathbb{E}[s(X) \mid A=a,Y=y] - \mathbb{E}[s(X) \mid Y=y]
    \label{eq:delta_out}  \\
    \Delta^{\phi}_{a,y,k}&=\mathbb{E}[\phi_{k}(X) \mid A=a,Y=y] - \mathbb{E}[\phi_{k}(X) \mid Y=y]
    \label{eq:delta_attr}
\end{align}
Substituting Eq.~\ref{eq:completeness} gives
\begin{equation}
    \Delta^{out}_{a,y}=\sum_{k=1}^{D}\Delta^{\phi}_{a,y,k}
    \label{eq:delta_relation}
\end{equation}
Hence, by the triangle inequality,
\begin{equation}
    \underbrace{| \Delta^{out}_{a,y} |}_{\ell^{out}_{a,y}} \leq \underbrace{\sum_{k=1}^{D}| \Delta^{\phi}_{a,y,k} |}_{\ell^{attr}_{a,y}}
    \label{eq:delta_ineq}
\end{equation}
A small output gap $\ell^{out}_{a,y}$ therefore does not imply a small attribution disparity $\ell^{attr}_{a,y}$ because positive and negative contribution differences may cancel during aggregation.

We characterize this mechanism through the cancellation ratio
\begin{equation}
    C_{a,y}=1-\frac{\ell^{out}_{a,y}}{\ell^{attr}_{a,y}} \qquad \ell^{attr}_{a,y} > 0
    \label{eq:cancellation}
\end{equation}
where $C_{a,y} \in [0,1]$. 
Larger values indicate that a greater fraction of absolute attribution disparity is masked through signed aggregation, whereas smaller values indicate that output-level gap relies less on cancellation.
We use the cancellation ratio as a diagnostic rather than a fairness criterion because low cancellation is desirable only when accompanied by small output and attribution disparities.

\subsection{Explanation-Aware Fairness Regularization}
\label{sssec:explain_reg}

Let
\begin{equation}
    \Omega=\{(a,y) \in \mathcal{A}\times\{0,1\}: P(A=a,Y=y)>0\}
    \label{eq:group_outcome}
\end{equation}
denote group-outcome combinations with nonzero support. We define the output- and attribution-level fairness objectives as
\begin{align}
    \mathcal{L}_{out}&=\frac{1}{|\Omega|} \sum_{(a,y) \in \Omega} |\Delta^{out}_{a,y}|
    \label{eq:out_loss} \\
    \mathcal{L}_{attr}&=\frac{1}{|\Omega|} \sum_{(a,y) \in \Omega} \sum_{k=1}^{D} |\Delta^{\phi}_{a,y,k}|
    \label{eq:attr_loss}
\end{align}
Averaging Eq.~\ref{eq:delta_ineq} gives $\mathcal{L}_{out} \leq \mathcal{L}_{attr}$.
Consistent with Eq.~\ref{eq:cancellation}, we define the aggregate cancellation ratio as
\begin{equation}
    C=1-\frac{\mathcal{L}_{out}}{\mathcal{L}_{attr}} \qquad \mathcal{L}_{attr} > 0
    \label{eq:cancellation_ratio}
\end{equation}

For optimization we use
\begin{equation}
    \bar{\mathcal{L}}_{attr}=\frac{\mathcal{L}_{attr}}{D}
    \label{eq:attr_loss_norm}
\end{equation}
which prevents the attribution penalty from scaling directly with representation dimensionality.

The proposed joint output-attribution objective simultaneously controls the observable group disparity and differences in feature contributions that may remain hidden in predictions.
With binary cross-entropy task loss $\mathcal{L}_{BCE}$, we optimize
\begin{equation}
    \mathcal{L}_{joint}=\mathcal{L}_{BCE}+\lambda_{out}\mathcal{L}_{out}+\lambda_{attr}\bar{\mathcal{L}}_{attr}
    \label{eq:joint_loss}
\end{equation}
where $out$ and $attr$ define output- and attribution-based fairness surrogates respectively, $\lambda_{out}$ and $\lambda_{attr}$ independently control these objective terms.
Output-only regularization corresponds to $\lambda_{attr}=0$ while attribution-only regularization is realized with $\lambda_{out}=0$, and these variants serve as direct ablations of the proposed joint formulation. 
Empirical risk minimization (ERM) corresponds to fairness-unaware optimization with $\lambda_{out}=0$ and $\lambda_{attr}=0$ whereas joint regularization requires $\lambda_{out} > 0$ and $\lambda_{attr} > 0$.
The attribution term constrains disparities in task-relevant predictive contributions whereas the output-level term constrains their final aggregate.
The joint objectives therefore seek output fairness supported by more closely aligned predictive contributions rather than parity arising primarily through cancellation.

\section{Experimental Setup}
\label{sec:setup}

\textbf{Datasets and tasks.}
We evaluate two binary medical image classification problems considering different imaging modalities and sensitive attributes.
HAM10000 \cite{tschandl2018ham10000} uses dermoscopic skin-lesion images for malignant/benign classification with gender as the sensitive attribute (Female and Male).
Harvard-GF \cite{luo2024harvard} uses retinal nerve fiber layer thickness (RNFLT) images for glaucoma detection with race as the sensitive attribute (Asian, Black/African American, and White/Caucasian).
HAM10000 uses five-fold lesion-level cross-validation (CV) and Harvard-GF uses its predefined train/validation/test split with five training seeds as $\{13,23,42,71,101\}$.
%
%We evaluate three binary medical image classification tasks with distinct sensitive attributes.
%On mBRSET \cite{wu2025portable}, the task is diabetic-retinopathy detection and sensitive attribute is gender.
%On HAM10000 \cite{tschandl2018ham10000}, we perform malignant/benign skin-lesion classification with gender as the sensitive attribute.
%On Harvard-GF \cite{luo2024harvard}, we classify glaucoma/healthy cases by using race as the sensitive attribute.
%For mBRSET and HAM10000, we use five-fold cross-validation with patient- and lesion-level splitting respectively.
%We use Harvard-GF with its predefined train/validation/test partition and five training seeds ($13,23,42,71,101$).
%

\begin{table}[t]
    \centering
    \scriptsize
    \setlength{\tabcolsep}{2.1pt}
    \begin{tabular}{lcc}
         \hline
         \textbf{Pair} & \textbf{HV: out$\rightarrow$joint.} &    $\mathbf{\Delta}$ \textbf{HV (+rel.) [95\% CI]} \\
         \hline
         \multicolumn{3}{l}{\textbf{HAM10000}} \\
         MeanEodd $\rightarrow$Joint$_1$   &   $0.2450\rightarrow 0.2622$ & $+0.0172$ (+7.02\%) [0.0072,0.0306]    \\
         Fairret $\rightarrow$Joint$_2$   &   $0.2502\rightarrow 0.2540$ & $+0.0038$ (+1.49\%) [0.0006,0.0255]    \\
         DiffEodd $\rightarrow$Joint$_3$   &   $0.2335\rightarrow 0.2581$ & $+0.0246$ (+10.57\%) [0.0175,0.0432]    \\
         \hline
         \multicolumn{3}{l}{\textbf{Harvard-GF}} \\
         MeanEodd $\rightarrow$Joint$_1$   &   $0.2474\rightarrow 0.2577$ & $+0.0103$ (+4.14\%) [0.0040,0.0293]    \\
         Fairret $\rightarrow$Joint$_2$   &   $0.2496\rightarrow 0.2552$ & $+0.0056$ (+2.24\%) [0.0008,0.0229]    \\
         DiffEodd $\rightarrow$Joint$_3$   &   $0.2528\rightarrow 0.2593$ & $+0.0065$ (+2.58\%) [0.0006,0.0206]    \\
         \hline
    \end{tabular}
    \caption{Fixed-$\tau=0.5$ BAcc-EOD frontier HV with reference $r=(0.5,1)$. Higher HV is better. Relative gain is $\Delta HV / HV_{\mathrm{out}}$. Confidence intervals are paired percentile 95\% CIs for $\Delta HV$. As baseline, HV scores of ERM are $0.2254$ and $0.2394$ for HAM10000 and Harvard-GF.}
    \label{tab:hv_results}
\end{table}

\textbf{Model and training.}
All methods use DINOv2-Small backbone \cite{oquab2023dinov2} with LoRA adaptation \cite{hu2022lora} ($r=8$, $\alpha=16$, dropout$=0$) and an explicit linear binary classification head.
Models are trained for 20 epochs with batch size 128 by using AdamW optimizer and cosine learning-rate scheduling.
The comparison includes the proposed explanation-aware fairness regularizer named AttrEodd alongside Empirical Risk Minimization (ERM), MeanEodd \cite{ozbulak2026lossconditioned}, Fairret \cite{buyl2024fairret}, DiffEodd \cite{han2024ffb}, and three attribution-augmented objectives as Joint$_1$: MeanEodd+AttrEodd, Joint$_2$: Fairret+AttrEodd, and Joint$_3$: DiffEodd+AttrEodd.
MeanEodd, Fairret, and DiffEodd are combined with BCE as output-level fairness-regularized objectives while the joint models additionally include the attribution-level term.
All named fairness objectives are differentiable proxies targeting the \eodd{} notion and we use shortened method names for compactness.
DiffEodd is extended for Harvard-GF by using the max-min group disparity for each outcome as the original study had the support only for binary sensitive groups.

By considering gradient scale of each method, we determined the $\lambda$ grids as
\begin{align}
    \lambda_{\mathrm{MeanEodd}} \in \{0.01,0.03,0.1,0.3,1\} \\
    \lambda_{\mathrm{AttrEodd}} \in \{1,3,10,30,100\} \\
    \lambda_{\mathrm{Fairret}} \in \{0.01,0.1,1,10,100\} \\
    \lambda_{\mathrm{DiffEodd}} \in \{0.2,0.5,1,2,4\}
\end{align}
Each joint family uses the Cartesian product of its output and attribution grids.
The study comprises a total of 960 trained models across five HAM10000 folds and five Harvard-GF training seeds.
%
%All methods use the same DINOv2-Small backbone \cite{oquab2023dinov2} with LoRA adaptation \cite{hu2022lora} ($r=8$, $\alpha=16$, dropout$=0$) and explicit linear binary classification head.
%Models are trained for 20 epochs with batch size 128 by using AdamW optimizer and cosine learning-rate scheduling.
%The comparison includes the proposed explanation-aware fairness regularizer named AttrEodd alongside Empirical Risk Minimization (ERM), MeanEodd \cite{ozbulak2026lossconditioned}, Fairret \cite{buyl2024fairret}, DiffEodd \cite{han2024ffb}, and three attribution-augmented variants as Joint\textsubscript{1}: MeanEodd+AttrEodd, Joint\textsubscript{2}: Fairret+AttrEodd, and Joint\textsubscript{3}: DiffEodd+AttrEodd.
%(GO: Mention here that every method is a proxy of EOdds fairness notion, this is not to include EOdds suffix into method names so that they do not occupy large space)
%
%By considering gradient scale of each method, we determined the frozen grids as $\lambda_{\mathrm{MeanEodd}} \in \{0.01,0.03,0.1,0.3,1\}$, $\lambda_{\mathrm{Attr}} \in \{1,3,10,30,100\}$, $\lambda_{\mathrm{Fairret}} \in \{0.01,0.1,1,10,100\}$, and $\lambda_{\mathrm{DiffEodd}} \in \{0.2,0.5,1,2,4\}$.
%Each joint family therefore contains the Cartesian product of its output and attribution level grids.
%

\textbf{Utility-fairness Pareto analysis.}
Every configuration is evaluated on held-out test data at a common threshold $\tau=0.5$ with no test-based hyper-parameter selection.
We compute equalized odds difference (EOD), balanced accuracy (BAcc), and area under the receiver operating characteristic curve (AUROC) and aggregate configurations across HAM10000 folds or Harvard-GF seeds.
For each method family, the non-dominated BAcc-EOD Pareto frontier is summarized by HV with reference point $r=(BAcc=0.5,EOD=1)$ using Fairical tool \cite{ozbulak_multi-objective_2025}.
We perform uncertainty estimation based on attribution-augmentation effects by using 10000 paired hierarchical bootstrap replicates.
Each replicate reconstructs the complete family frontier while resampling lesions within HAM10000 folds or held-out samples together with training-seed variation for Harvard-GF.
We report $\Delta HV=HV_{\mathrm{joint}}-HV_{\mathrm{out}}$ with percentile 95\% confidence intervals.
%
%Our main analysis evaluates every trained configuration on its held-out test split using the same fixed decision threshold $\tau=0.5$ and no hyper-parameter or threshold is selected by using test performance.
%For each configuration, we compute equalized odds difference (EOD), balanced accuracy (BAcc), and area-under-the curve (AUROC).
%Metrics are aggregated across the five folds for mBRSET/HAM10000 and across the five training seeds for Harvard-GF.
%
%For each method family, we construct the non-dominated BAcc-EOD frontier and summarize it using hypervolume (HV) with reference point ($r=(BAcc=0.5,EOD=1)$).
%The BAcc coordinate corresponds to the non-informative baseline for balanced binary classification such that HV rewards utility-fairness improvements only in the above-change region.
%
%Uncertainty in family-level performance is estimated with $10000$ paired hierarchical bootstrap replicates.
%The complete method-specific Pareto frontier is reconstructed with every replicate.
%Resampling preserves the corresponding held-out unit structure: patient within fold for mBRSET, lesion within fold for HAM10000, and test sample together with training-seed variation for Harvard-GF.
%We report paired $\Delta$HV with percentile 95\% confidence intervals (CIs).

\textbf{\emph{A posteriori} analysis.}
We additionally characterize whether each trained family contains a low-disparity operating point while preserving utility.
For each HAM10000 fold, the test configuration and threshold minimizing EOD are selected subject to
\begin{equation}
    BAcc \geq BAcc_{\mathrm{ERM}}^{\max}-0.01
    \label{eq:op_rule}
\end{equation}
For Harvard-GF, one common configuration and threshold are selected from mean performance across the five seeds.
This analysis measures attainable operating-point coverage for each method and compares them.

\begin{figure}[t]
\begin{minipage}[b]{0.49\linewidth}
  \centering
  \centerline{\includegraphics[width=4.0cm]{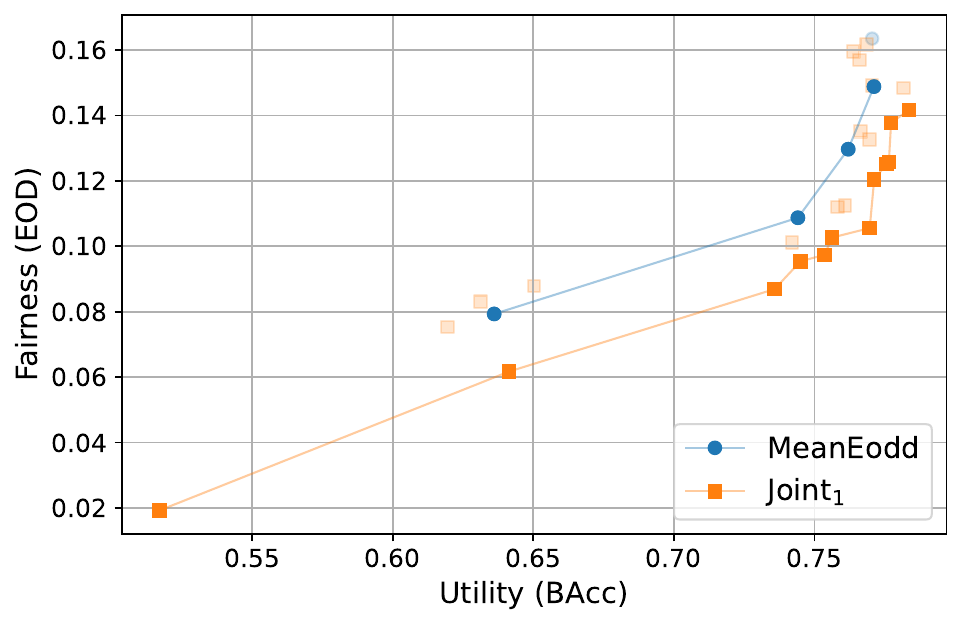}}
  \centerline{(a) HAM10000}\medskip
\end{minipage}
\begin{minipage}[b]{0.49\linewidth}
  \centering
  \centerline{\includegraphics[width=4.0cm]{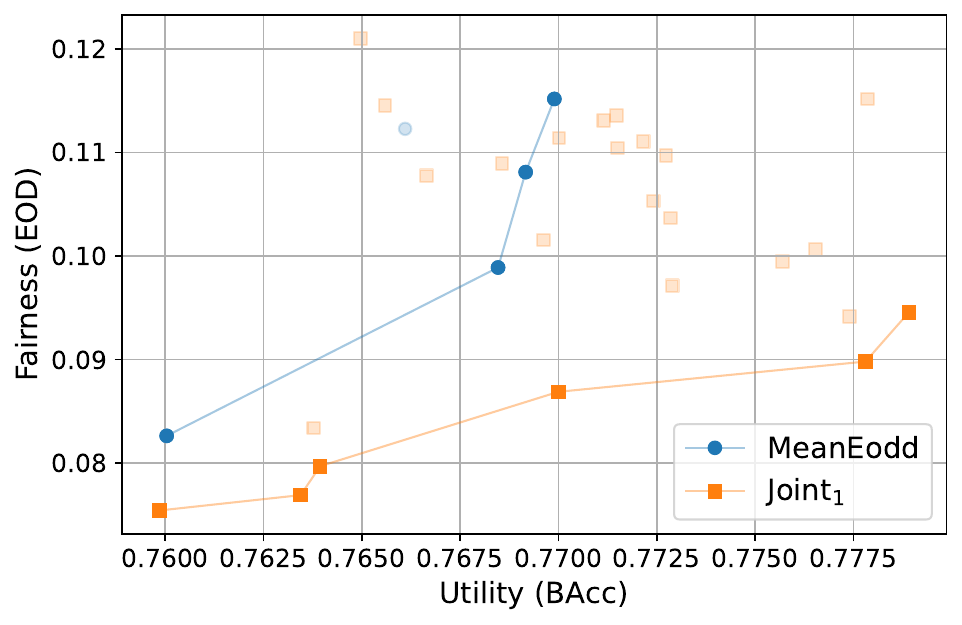}}
  \centerline{(b) Harvard-GF}\medskip
\end{minipage}
\caption{BAcc-EOD Pareto frontiers for MeanEodd and Joint$_1$ at $\tau=0.5$. Each point corresponds to a trained hyper-parameter configuration. Curves connect only non-dominated solutions and bottom-right indicates favorable trade-offs with higher BAcc and lower EOD. Attribution augmentation increases HV on both datasets.}
\label{fig:pareto_results}
\end{figure}

\begin{table*}[t]
\centering
    \scriptsize
    \setlength{\tabcolsep}{2.6pt}
    \begin{tabular}{llccccc}
    \hline
    \textbf{Dataset} & \textbf{Pair} & \textbf{EOD ($\downarrow$)} & \textbf{BAcc ($\Delta$)} & \textbf{AUROC ($\Delta$)} & $\mathbf{\mathcal{L}_{attr} (\downarrow)}$ & $\mathbf{C (\Delta)}$ \\
    \hline
    \textbf{HAM10000} & MeanEodd $\rightarrow$ Joint$_1$ & $5.52 \rightarrow 4.20$ (23.9\%) & $82.34 \rightarrow   82.20$ ($-0.14$) & $91.08 \rightarrow 91.14$ ($+0.06$) & $0.427 \rightarrow 0.178$ (58.2\%) & $0.735 \rightarrow 0.347$ ($-0.388$) \\
     & Fairret $\rightarrow$ Joint$_2$ & $5.90 \rightarrow 3.79$ (35.8\%) & $82.40 \rightarrow 82.00$ ($-0.40$) & $90.84 \rightarrow 90.98$ ($+0.14$) & $0.457 \rightarrow 0.263$ (42.5\%) & $0.631 \rightarrow 0.446$ ($-0.185$) \\
     & DiffEodd $\rightarrow$ Joint$_3$ & $5.99 \rightarrow 3.67$ (38.7\%) & $82.05 \rightarrow 82.09$ ($+0.04$) & $90.33 \rightarrow 91.00$ ($+0.67$) & $0.376 \rightarrow 0.216$ (42.7\%) & $0.775 \rightarrow 0.440$ ($-0.335$) \\
     \hline
     \textbf{Harvard-GF} & MeanEodd $\rightarrow$ Joint$_1$ & $8.30 \rightarrow 6.09$ (26.5\%) & $77.13 \rightarrow 77.51$ ($+0.38$) & $84.96 \rightarrow 85.06$ ($+0.10$) & $0.298 \rightarrow 0.054$ (82.0\%) & $0.732 \rightarrow 0.460$ ($-0.272$) \\
     & Fairret $\rightarrow$ Joint$_2$ & $8.34 \rightarrow 7.17$ (14.0\%) & $78.02 \rightarrow 78.32$ ($+0.30$) & $85.26 \rightarrow 85.28$ ($+0.02$) & $0.292 \rightarrow 0.098$ (66.5\%) & $0.680 \rightarrow 0.291$ ($-0.389$) \\
     & DiffEodd $\rightarrow$ Joint$_3$ & $7.20 \rightarrow 6.55$ (9.0\%) & $77.36 \rightarrow 77.58$ ($+0.22$) & $84.98 \rightarrow 85.01$ ($+0.03$) & $0.266 \rightarrow 0.049$ (81.7\%) & $0.825 \rightarrow 0.438$ ($-0.387$) \\
     \hline    
    \end{tabular}
    \caption{\emph{A posteriori} utility-constrained operating points. EOD, BAcc, and AUROC are reported in percentage and parentheses give relative EOD/$\mathcal{L}_{attr}$ reductions or absolute percentage/ratio changes. $\mathcal{L}_{attr}$ is the unnormalized attribution-disparity from Eq.~\ref{eq:attr_loss} and optimization uses $\bar{\mathcal{L}}_{attr}$.}
    \label{tab:aposteriori_results}
\end{table*}

\section{Results and Analysis}
\label{sec:results}

\textbf{Attribution augmentation improves the attainable utility-fairness trade-off.}
All six comparisons showed gains in $\Delta HV$ with paired 95\% CIs consistently supporting the joint formulations (Table~\ref{tab:hv_results}).
On HAM10000, Joint$_1$, Joint$_2$, and Joint$_3$ improved HV over MeanEodd, Fairret, and DiffEodd by $7.02\%$, $1.49\%$, and $10.57\%$ respectively.
Corresponding gains on Harvard-GF were $4.14\%$, $2.24\%$, and $2.58\%$.
Fig.~\ref{fig:pareto_results} illustrates the corresponding BAcc-EOD Pareto frontiers for MeanEodd and Joint$_1$ on both datasets.
Thus our hypothesis is supported across both datasets and all three output-level \eodd{} surrogates.
ERM additionally provides a utility-only reference for the attainable trade-off. Its HV was $0.2254$ on HAM10000 and $0.2394$ on Harvard-GF, and every fairness-aware regularizer family attained a higher score on both datasets.
Relative to ERM, Joint$_1$, Joint$_2$, and Joint$_3$ increased HV by $16.3\%$, $12.7\%$, and $14.5\%$ on HAM10000, and by $7.6\%$, $6.6\%$, and $8.3\%$ on Harvard-GF respectively.

%
%\textbf{Primary utility-fairness trade-off.}
%Among the six model families, Joint\textsubscript{1} achieved the largest fixed-threshold HV on all three datasets (Table~\ref{tab:hv_results}).
%Relative to MeanEodd, Joint\textsubscript{1} increased HV by $2.6\%$ on mBRSET, $7.0\%$ on HAM10000, and $4.1\%$ on Harvard-GF.
%The corresponding paired HV differences were positive on all three datasets with 95\% confidence intervals as $+0.0068$ [$+0.0012,+0.0293$], $+0.0172$ [$+0.0071,+0.0306$], and $+0.0103$ [$+0.0037,+0.0296$] respectively.
%Relative to ERM, the HV gains were approximately $2.9\%$, $16.3\%$, and $7.6\%$.
%
%The additional Joint\textsubscript{2} and Joint\textsubscript{3} experiments test whether this benefit extends beyond Joint\textsubscript{1} surrogate.
%5Their principal comparisons are therefore $HV_{\mathrm{Joint_{2}}}-HV_{\mathrm{Fairret}} \quad \text{and} \quad HV_{\mathrm{Joint_{3}}}-HV_{\mathrm{DiffEodd}}$ using exactly the same fixed-threshold and bootstrap protocol.

\textbf{\emph{A posteriori} operating points support the attribution mechanism.}
Under the predefined utility-constraint in Eq.~\ref{eq:op_rule}. every augmented family attained lower EOD than its corresponding output-only counterpart (Table~\ref{tab:aposteriori_results}).
Relative EOD reductions ranged from $9.0\%$ to $38.7\%$ while BAcc changed at most $0.40\%$. AUROC was maintained or slightly improved in all six comparisons.
At the same operating points, $\mathcal{L}_{attr}$ decreased by $42.5\%$-$82.0\%$ and cancellation reduced in all six comparisons.
Thus the family-level trade-off improvements are accompanied by operating points with lower feature-level disparity and less dependence on signed cancellation.
%
%This complementary analysis provides a consistent operating point level view of the same phenomenon.
%For MeanEodd vs. Output+Attr, EOD decreased from $2.15$ to $1.13$ on mBRSET, from $5.52$ to $4.20$ on HAM10000, and from $8.30$ to $6.09$ on Harvard-GF, corresponding to relative improvements of $47.44\%$, $23.91\%$, and $26.62\%$.
%These performances were obtained while preserving predictive utility: BAcc changed from $79.88$ to $80.08$, $82.34$ to $82.20$, and $77.13$ to $77.51$ respectively, while AUROC remained comparable or slightly increased.
%(GO: if space permits, we should put table for more details)
%Similarly, relative EOD improvement of Output+Attr for ERM performances as $4.04$, $13.36$, and $10.46$ is $72.02\%$, $68.56\%$, and $41.77\%$ respectively.
%(GO: is this ERM comparison good? should we also share BAcc-AUROC situation?)
%
% GO: Add new experiments for Fairret/Fairret+Attr and DiffEodd/DiffEodd+Attr and report resulting EOD, BAcc, AUROC pairs
%
%\textbf{Attribution-level mechanism.}
%Output parity alone can arise despite substantial canceling attribution differences.
%Consistent with the proposed mechanism, the completed Output+Attr experiments reduce attribution disparity and cancellation relative to Output surrogate while simultaneously improving the BAcc-EOD trade-off frontier.
%We therefore treat $\mathcal{L}_{attr}$ and cancellation ratio $C$ as diagnostic rather than standalone fairness notion.
%
%GO: test whether Fairret+Attr and DiffEodd+attr similarly reduce L_attr and C relative to their corresponding Output

\section{Discussion and Conclusion}
\label{sec:discussion}

The experiments provide two complementary levels of evidence that feature-level attribution disparity contains information that is not controlled by output fairness alone.
At the family level, adding the identical $\bar{\mathcal{L}}_{attr}$ objective to MeanEodd, Fairret, and DiffEodd improves held-out BAcc-EOD HV on both datasets with all six paired 95\% confidence intervals as entirely positive.
The effect is therefore not specific to the geometry of any of output fairness based surrogates as attribution regularization consistently improves the attainable utility-fairness trade-offs across three different output-level \eodd{} objectives.

The \emph{a posteriori} analysis provides a complementary view as every augmented family contains a lower EOD operating point at comparable utility while consistently exhibiting lower $\mathcal{L}_{attr}$ and lower cancellation $C$.
This analysis demonstrates that the improved method families contain practically favorable low-disparity solutions and that these solutions are supported by more closely aligned predictive feature contributions.

Our observations are based on empirical held-out Pareto frontiers over the prespecified regularization grids.
In this work, the exact output-attribution decomposition relies on the linear classification head, and its extension to nonlinear prediction heads requires further study.
Evaluation is also limited to two binary medical imaging tasks although they differ in imaging modality and sensitive attribute.

In conclusion, output parity alone does not determine how predictive evidence is distributed across demographic groups.
Explanation-aware attribution regularization complements output-level fairness by constraining disparities in the task-relevant feature contributions producing the decision.
Across several existing regularization approaches for fairness, this additional constraint consistently improves the attainable utility-fairness trade-off while providing operating points with lower attribution disparity and less cancellation.

\section{Compliance with ethical standards}
\label{sec:ethics}

This research study was conducted retrospectively using human subject data made available in open access through the HAM10000 and Harvard-GF datasets. 
Ethical approval was not required as confirmed by the license attached with the open access data.
%GO: fill source details

\section{Acknowledgments}
\label{sec:acknowledgments}

This work was supported by the Swiss National Science Foundation (SNSF) through the project FairMI - Machine Learning Fairness with Application to Medical Images under grant number 214653. 
We also thank Fundação de Amparo à Pesquisa do Estado de São Paulo (FAPESP), grant 21/14725-3.
The authors declare no relevant financial or non-financial conflicts of interest.

\bibliographystyle{IEEEbib}
\bibliography{strings,refs}

@article{ozbulak_multi-objective_2025,
   title={A Multi-Objective Evaluation Framework for Analyzing Utility-Fairness Trade-Offs in Machine Learning Systems},
   author={Özbulak, Gökhan and Jimenez-del-Toro, Oscar and Fatoretto, Maíra and Berton, Lilian and Anjos, André},
   journal={Machine Learning for Biomedical Imaging},
   volume={3},
   number={Special issue on FAIMI},
   pages={938--957},
   doi={10.59275/j.melba.2025-ab9a},
   year={2025}
}

@inproceedings{han2024ffb,
    title={{FFB}: A Fair Fairness Benchmark for In-Processing Group Fairness Methods},
    author={Xiaotian Han and Jianfeng Chi and Yu Chen and Qifan Wang and Han Zhao and Na Zou and Xia Hu},
    booktitle={The Twelfth International Conference on Learning Representations},
    year={2024},
    url={https://openreview.net/forum?id=TzAJbTClAz}
}

@inproceedings{buyl2024fairret,
    title={fairret: a Framework for Differentiable Fairness Regularization Terms},
    author={Maarten Buyl and MaryBeth Defrance and Tijl De Bie},
    booktitle={The Twelfth International Conference on Learning Representations},
    year={2024},
    url={https://openreview.net/forum?id=NnyD0Rjx2B}
}

@inproceedings{ozbulak2026lossconditioned,
    title={Loss-Conditioned Utility-Fairness Boundary Modeling for Medical Imaging},
    author={G{\"o}khan {\"O}zbulak and Oscar Jimenez-del-Toro and Lilian Berton and Andre Anjos},
    booktitle={Joint FAIMI, BRIDGE, and EPIMI Workshop at MICCAI 2026},
    year={2026},
    url={https://openreview.net/forum?id=4wSEYsqRXm}
}

@inproceedings{hu2022lora,
    title={Lo{RA}: Low-Rank Adaptation of Large Language Models},
    author={Edward J Hu and yelong shen and Phillip Wallis and Zeyuan Allen-Zhu and Yuanzhi Li and Shean Wang and Lu Wang and Weizhu Chen},
    booktitle={International Conference on Learning Representations},
    year={2022},
    url={https://openreview.net/forum?id=nZeVKeeFYf9}
}

@article{oquab2023dinov2,
  title={Dinov2: Learning robust visual features without supervision},
  author={Oquab, Maxime and Darcet, Timoth{\'e}e and Moutakanni, Th{\'e}o and Vo, Huy and Szafraniec, Marc and Khalidov, Vasil and Fernandez, Pierre and Haziza, Daniel and Massa, Francisco and El-Nouby, Alaaeldin and others},
  journal={arXiv preprint arXiv:2304.07193},
  year={2023}
}

@article{luo2024harvard,
  title={Harvard glaucoma fairness: a retinal nerve disease dataset for fairness learning and fair identity normalization},
  author={Luo, Yan and Tian, Yu and Shi, Min and Pasquale, Louis R and Shen, Lucy Q and Zebardast, Nazlee and Elze, Tobias and Wang, Mengyu},
  journal={IEEE Transactions on Medical Imaging},
  volume={43},
  number={7},
  pages={2623--2633},
  year={2024},
  publisher={IEEE}
}

@article{tschandl2018ham10000,
  title={The HAM10000 dataset, a large collection of multi-source dermatoscopic images of common pigmented skin lesions},
  author={Tschandl, Philipp and Rosendahl, Cliff and Kittler, Harald},
  journal={Scientific data},
  volume={5},
  number={1},
  pages={180161},
  year={2018},
  publisher={Nature Publishing Group}
}

@article{larrazabal2020gender,
  title={Gender imbalance in medical imaging datasets produces biased classifiers for computer-aided diagnosis},
  author={Larrazabal, Agostina J and Nieto, Nicol{\'a}s and Peterson, Victoria and Milone, Diego H and Ferrante, Enzo},
  journal={Proceedings of the National Academy of Sciences},
  volume={117},
  number={23},
  pages={12592--12594},
  year={2020},
  publisher={National Academy of Sciences}
}

@article{seyyed2021underdiagnosis,
  title={Underdiagnosis bias of artificial intelligence algorithms applied to chest radiographs in under-served patient populations},
  author={Seyyed-Kalantari, Laleh and Zhang, Haoran and McDermott, Matthew BA and Chen, Irene Y and Ghassemi, Marzyeh},
  journal={Nature medicine},
  volume={27},
  number={12},
  pages={2176--2182},
  year={2021},
  publisher={Nature Publishing Group US, New York}
}

@article{yang2024limits,
  title={The limits of fair medical imaging AI in real-world generalization},
  author={Yang, Yuzhe and Zhang, Haoran and Gichoya, Judy W and Katabi, Dina and Ghassemi, Marzyeh},
  journal={Nature medicine},
  volume={30},
  number={10},
  pages={2838--2848},
  year={2024},
  publisher={Nature Publishing Group US, New York}
}

@inproceedings{lundberg2020explaining,
  title={Explaining quantitative measures of fairness},
  author={Lundberg, Scott M},
  booktitle={Fair \& Responsible AI Workshop@ CHI2020},
  year={2020}
}

@article{hardt2016equality,
  title={Equality of opportunity in supervised learning},
  author={Hardt, Moritz and Price, Eric and Srebro, Nati},
  journal={Advances in neural information processing systems},
  volume={29},
  year={2016}
}

@article{wang2026procedural,
  title={Procedural fairness in machine learning},
  author={Wang, Ziming and Huang, Changwu and Tang, Ke and Yao, Xin},
  journal={Journal of Artificial Intelligence Research},
  volume={85},
  year={2026}
}

@article{shahbazian2025joint,
  title={Joint explainable and fair AI in healthcare},
  author={Shahbazian, Reza and Trubitsyna, Irina and Mirtaheri, Seyedeh Leili},
  journal={BMC Medical Informatics and Decision Making},
  volume={25},
  number={1},
  pages={403},
  year={2025},
  publisher={Springer}
}

\end{document}